\documentclass[11pt]{article}

\usepackage[a4paper, margin=1in]{geometry}

\usepackage[round]{natbib}

\usepackage{comment}
\usepackage{amsfonts}       
\usepackage{nicefrac}       

\usepackage{algorithm}
\usepackage{algpseudocode}
\usepackage{graphicx}
\usepackage{enumitem}
\usepackage{booktabs}

\usepackage{amsmath}
\usepackage{amsthm}

\usepackage{subcaption} 

\begin{document}

\title{Characterization of Transfer Using Multi-task Learning Curves}

\author{ 
András Millinghoffer \and Bence Bolgár \and Péter Antal\thanks{Corresponding author: \texttt{antal@mit.bme.hu}} \\
Department of Artificial Intelligence and Systems Engineering\\
Faculty of Electrical Engineering and Informatics\\
Budapest University of Technology and Economics\\
Műegyetem rkp. 3, H-1111 Budapest, Hungary
}
\maketitle

\begin{abstract}
Transfer effects manifest themselves both during training using a fixed data set and in inductive inference using accumulating data. We hypothesize that perturbing the data set by including more samples, instead of perturbing the model by gradient updates, provides a complementary and more fundamental characterization of transfer effects. To capture this phenomenon, we quantitatively model transfer effects using multi-task learning curves approximating the inductive performance over varying sample sizes. We describe an efficient method to approximate multi-task learning curves analogous to the Task Affinity Grouping method applied during training. We compare the statistical and computational approaches to transfer, which indicates considerably higher compute costs for the previous but better power and broader applicability. Evaluations are performed using a benchmark drug-target interaction data set. Our results show that learning curves can better capture the effects of multi-task learning and their multi-task extensions can delineate pairwise and contextual transfer effects in foundation models.  
\end{abstract}

\section{Introduction}\label{s:intro}

Transfer effects between machine learning tasks are of central importance, indicated by their sustained investigation in many subfields, such as in multi-task learning (MTL)~\citep{caruana1997multitask,xin2022current}, in transfer learning and model transformation/distillation~\citep{neal2001transferring,antal2003bayesian,zhang2022survey}, in multimodal fusion~\citep{antal2004using,baltruvsaitis2018multimodal,xu2023multimodal}, in multi-objective optimization~\citep{sener2018multi}, in curriculum learning~\citep{bengio2009curriculum}, in learning with prior knowledge and few-shot learning~\citep{song2023comprehensive}, in meta-learning~\citep{vilalta2002perspective,klein2016learning,hospedales2021meta}, and in active learning and adaptive study design~\citep{tong2001active,rzhetsky2015choosing,mahmood2022optimizing}. Despite a wide variety of efforts, questions are still surmounting in multi-task learning scenarios about its useful or even negative effects~\citep{xin2022current,zhang2022survey}. On the contrary, transformer-based pre-trained large language models and multimodal architectures, a.k.a. foundation models, achieved considerable progress in unsupervised learning of unified representations and utilizing them in domain-wise and cross-domain tasks, even showing signs of artificial general intelligence~\citep{subramanian2018learning,brown2020language,bubeck2023sparks,taylor2022galactica}. Although early works on MTL emphasized inductive bias and theoretical models for performance bounds on MTL were developed~\citep{caruana1997multitask,baxter2000model,ben2003exploiting,ben2010theory}, many recent works concentrated on training, e.g., on orchestrating training~\citet{chen2018gradnorm} and utilizing cross-training linkage between tasks to design task decomposition~\citet{standley2020tasks,fifty2021efficiently,song2022efficient}.
\begin{figure*}[!h]
\centering
\vspace{.3in}
\includegraphics[width=0.875\textwidth, keepaspectratio] {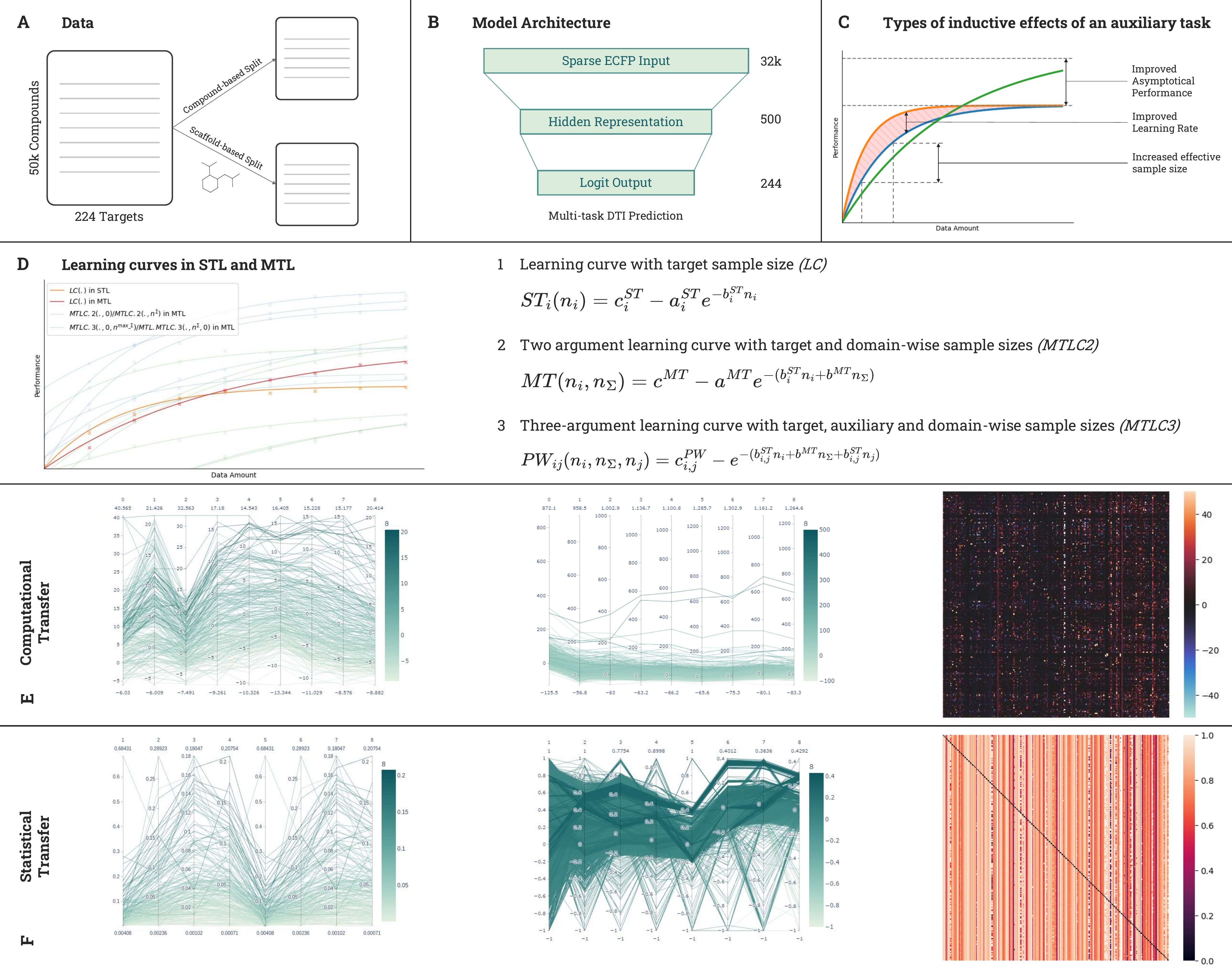}
\vspace{.3in}
\caption{\textbf{Overview.} \textbf{(A) Data:} The main data set contains 50k samples with 32k complete input descriptors and 244 output tasks, which is highly incomplete (90\%< is missing). Cross-validation folds are generated at the sample and perturbed sample group (scaffold) levels. \textbf{(B) Model:} The 244 tasks correspond to 244 head nodes on a shared foundation layer with 2k nodes in a multiple-output MLP. \textbf{(C) Transfer:} Transfer learning can change the effective sample size, learning rate, and asymptotic limit of a target task. \textbf{(D) Learning curves:} The effects of sample sizes for the target (horizontal axis), the auxiliary (illustrated at 0 and $n^\mathrm{max}_\mathrm{aux}$) and the complementary (illustrated at 0 and $n^\mathrm{max}_{\Sigma}$) tasks are modeled in both single and multi-task learning (STL/MTL) scenarios using 1/2/3-argument learning curves (LC, MTLC.2/3). \textbf{(D-E) Computational transfer:} Transfer effects using cross-task training versus cross-task samples are analyzed and compared for systematically varying sample sizes.}
\end{figure*}\label{fig:overview}

In this paper, we also utilize the dynamics of learning to characterize transfer effects, but as a function of sample size instead of as a function of training cycle/time using a fixed-sized data set. We hypothesize that the more constrained dynamics of statistical learning over data allow a more fundamental characterization of multi-task transfers than in the more engineered training process. We systematically compare the corresponding dual statistical and computational aspects of transfer effects, where the basis of our investigation is the transfer effect of an extra block of training data versus the transfer effect of an extra training cycle. In line with other approaches, we perform this comparison in a pairwise manner, investigating the effect of an auxiliary task on a target task~\citet{standley2020tasks,fifty2021efficiently,song2022efficient}, but we also compare the domain-wide transfer effects, investigating the transfer effect of all the other tasks as an auxiliary set. This domain-wide scenario provides a model for the transfer effects in foundation models, and we try to delineate domain-wide and pairwise transfer effects by using parametric learning curves (LCs)~\citep{viering2022shape}. Respectively, we propose novel methods to investigate parametric forms for multi-task learning curves (MTLCs) explicitly modeling transfer effects of domain-wide tasks as context and pairwise effects between tasks. We demonstrate our method using real-world drug-target interaction (DTI) benchmarks~\citep{tang2014making,wang2022profiling}.

The structure of the paper is as follows. First, we motivate and discuss properties of the proposed multi-task learning curves. Second, we describe an efficient estimation method for these learning curves. Third, we perform a systematic evaluation of MTLCs and comparison with existing methods using a benchmark DTI data set, specifically, we show that specialties of DTI data sets render both gradient surgery~\citet{chen2018gradnorm} and task decomposition methods inefficient~\citet{fifty2021efficiently,song2022efficient}. We systematically investigate the behavior of TAG as a function of sample size~\citet{fifty2021efficiently} and demonstrate that LCs can better detect effects of multi-task learning and the extended MTLCs can delineate different types of transfer effects. We also show that MTLCs allow a robust prediction of performance gain from further data for a wide range of tasks and sample sizes, suggesting its use in active learning for the DTI domain~\citet{mahmood2022optimizing}. Fig.~\ref{fig:overview} summarizes the background of our approach, the proposed multi-task learning curves, and its investigation.

\section{Related works}\label{s:related}

Inductive bias, transfer effects between tasks, was a central concept in \citet{caruana1997multitask}'s seminal work and series of theoretical works investigated statistical learning theory of transfer learning~\citet{baxter2000model,ben2003exploiting,ben2010theory}. However, theoretical analysis of transfer effects in the case of MTL with hard sharing hidden layer(s) is still an active area of research~\citet{obozinski2010joint,pentina2017multi,wu2020understanding,galanti2022improved}. Despite the wide range of techniques that were proposed for MTL~\citep{chen2018gradnorm,sener2018multi,yu2020gradient,liu2019loss,meng2021multi}, systematic empirical evaluations indicate very modest or not significant advance~\citet{xin2022current}, especially, that in many applications significant negative transfers also emerge~\cite{zhang2022survey}. A targeted solution to avoid negative transfer would be an efficient screening for it and a decomposition of tasks accordingly~\cite{standley2020tasks,fifty2021efficiently,song2022efficient,jiang2023forkmerge}.

In the life sciences, the drug-target interaction prediction problem has long been viewed as the ideal case for MTL~\citet{jacob2008virtual,rosenbaum2013inferring,unterthiner2014multi,cai2020transfer}. However, results are still mixed and negative transfer is persistent~\citet{li2019deep,moon2022prediction,zhou2021multidti,valsecchi2022predicting,lin2022generalizeddta,wang2022profiling,heyndrickx2023conformal}. Factors behind the much-awaited transfer effects, especially stemming from the shared foundation feature layers/maps, are still not known and currently available hypotheses for transfer were formulated in small-scale investigations using only a dozen targets (tasks)~\cite{xu2017demystifying}. 

The high incompleteness of real-world, large-scale DTI data sets seems to be a major obstacle to the adoption of general multi-task learning theories and models, and as this sparsity and not missing-at-random properties are inherent in many discovery problems, this drove our work towards novel solutions.

\section{Multi-task learning curves}\label{s:MTLCs}

The term learning curves is equally used in the literature to describe both the empirical model performance or error in the training process and the characteristics of the generalization performance of a learning algorithm with increasing amounts of data, for an explicit modeling of learning curves in training, see e.g., \citet{klein2016learning}. Following~\citet{viering2022shape}, we refer to the earlier as the training curve, and for the latter, we define the learning curve as follows. We assume that a learning algorithm $\mathcal{A}$ selects a multi-task binary decision function $f$  $\mathbb{R}^{d}\to \{0,1\}^{K}$ from a function class $\mathcal{F}$, where $d$ denotes the input dimension and $K$ denotes the number of tasks. The input and output spaces are denoted by $\mathcal{X}$ and $\mathcal{Y}$, and their joint probability distribution by $P_{XY}$.  Following the hard parameter sharing approach, the function $f$ is defined as a composite mapping
\begin{equation}\label{eq:CompositeMLP}
    f = g \circ h,
\end{equation}
where 
\begin{equation}\label{eq:TrunkMLP}
g \in \mathcal{G} \subset \{ g': \mathbb{R}^{d}\to \mathbb{R}^{r} \},
\end{equation}
is the shared feature map and $h$ denotes all the classifiers
\begin{equation}\label{eq:HeadMLP}
h=\{h_i \in \mathcal{H} \subset \{ h': \mathbb{R}^{r} \to \{0,1\}\}\}_{i=1}^K,
\end{equation}

for each task $i=1,\ldots K$~\citep{galanti2022improved}. The shared feature map $g$ is frequently called trunk, which provides a common, universal representation for all the predictor functions $h_i$ in a domain analogously to pretrained transformer-based foundation models~\citep{zhou2023comprehensive}.

We assume a loss function $L(\hat{y},y)$ specifying the error for the prediction $\hat{y}=f(x)$ and true value $y$, where $\hat{y},y \in \{0,1\}^{K}$. This allows the definition of an idealized performance measure for a given function $f$ and function class $\mathcal{F}$
\begin{equation}
    L(f)=\mathbb{E}_{P_{XY}} \left[ L(f(X),Y)\right],\,\,L^*_\mathcal{F}=\inf_{f \in \mathcal{F}} L(f),
\end{equation}

where the expected loss or risk $L(f)$ denotes the asymptotic generalization performance of $f$ given an unlimited amount of i.i.d. samples drawn from $P_{XY}$ and $L^*_\mathcal{F}$ denotes the asymptotic limit for an idealized learning algorithm using this idealized data scenario. In practice, the learning algorithm $\mathcal{A}$ uses $n$ samples as the training data set $D_n$ to select an empirically optimal model $\hat{f}$ and its risk is estimated using a test data set
\begin{equation}\label{eq:EstimatedRisk}
    \hat{L}(\hat{f}),\mathrm{where }\hat{f}=\mathcal{A}(D_N).
\end{equation}

For a given ordering $\prec$ of the samples in data set $D_N$, we can define a learning curve as follows
\begin{equation}\label{eq:EmpiricalLearningCurve}
    \hat{L}^\prec_n(\mathcal{A}(D_n)),
\end{equation}
where $D_n$ denotes first $n$ samples w.r.t. $\prec$. Following \citet{viering2022shape}, taking expectations over the data set $D_n$ results in an idealized learning curve
\begin{equation}\label{eq:IdealLearningCurve}
    \Bar{R}_n(\mathcal{A})=\mathbb{E}_{P_{D_n}} \left[ \hat{L}_n(\mathcal{A}(D_n))\right].
\end{equation}

The multi-task loss function usually can be decomposed for the tasks
\begin{equation}
    L(\hat{y},y) = \sum_{i=1}^K L_i(\hat{y_i},y_i).
\end{equation}

In the case of loss-free performance metrics, such as the Area Under the Receiver Operating Curve (AUROC) and the Area Under the Precision-Recall Curve (AUPR), the performance of a multi-task scoring function $f$  $\mathbb{R}^{d}\to \mathbb{R}^{K}$ can be estimated analogously using the predicted scores $\hat{y}_{i,.}=\hat{y}_i^{(l)}$ for task $i$ and samples $l=1\ldots n$ and the outcomes $y_{i,.}$
\begin{equation}
    \hat{L}^\prec_n(\mathcal{A}(D_n))=\sum_{i=1}^K L_i(\hat{y_{i,.}},y_{i,.}).
\end{equation}

Finally, we expect complete $x$ inputs, but in the case of high-dimensional outcomes, outcomes are frequently incomplete. In this case, $\hat{L}^\prec_n(\mathcal{A}(D_n))$ denotes the sum of the respective estimated losses based on the $n_i$ number of outcomes for $Y_i$ present in $D_n$. 

Our central assumption is that learning curves show strong regularities compared to the chaotic behavior of training curves, which regularities can be exploited to improve the detection and characterization of multi-task learning effects, especially transfer effects. A notable result shows that for a target task $T$, the generalization error is bounded as follows (for conditions and details see Theorem~4~\citet{ben2010theory}):
\begin{align}
    L_{P,T}(\hat{f})\le L_{P,T}(f_T^*)+2\sqrt{\frac{2D\log(2(n+1))
    +2\log(\frac{4}{\delta})}{n}}\nonumber\\+\frac{2}{K}\sum_{i=1}^K \min_{f \in \mathcal{F}}\{L_{P,T}(f)+L_{P,i}(F)\}, \label{eq:MTSampleSum}
\end{align}

where $D$ is the VC dimension of $\mathcal{F}$, $\hat{f} \in \mathcal{F}$ is the empirical minimizer for the tasks, $f_T^*=\min_{f \in \mathcal{F}}L_{P,T}(f)$ is the target error minimizer. Notably, $n=\sum_i n_i$ denotes the sum of the sample sizes for all the tasks and the last error term expresses the penalty for the discrepancies between the tasks and the target task $T$. For results on bounds using AUROC, see e.g., ~\citep{usunier2005data}, for extensions to the case of foundation models using hard parameter sharing between the tasks, see~\citet{wu2020understanding,galanti2022improved}.

Although theoretical bounds for single task, multi-task, and foundation models support rational choices for the analytical forms of the learning curves, there are further important factors in selecting: (1) reliable estimation of risk in Eq.~\ref{eq:EstimatedRisk} is not possible for low data tasks in case of highly incomplete multi-task data sets as the number of outcomes can be very low in the test data, (2) highly incomplete data also excludes the estimation of task covariances, (3) non-standard performance measures, which are motivated by the discovery nature of the problem, class imbalances, incompleteness of the data, can be efficiently computed, but may have no tight bounds, e.g., enrichment methods with reject and ranking options, (4) interpretability of learning curve parameters support richer options for multi-task extension and characterization of transfer effects, and (5) efficient estimation of the learning curve, especially its multitask extension.

The repertoire of LC approximations is very broad~\citet{viering2022shape,mohr2022learning}. Based on interpretability and applicability for performance measures such as AUROC and AUPR, the following learning curve approximations seemed to be the best choices:
\begin{align}
    \mathrm{EXP4}(n) &= c - \mathrm{exp}(b - a n^{\alpha}),\label{eq:exp4}\\
    \mathrm{EXP3}(n) &= c - \mathrm{exp}(b - a n),\label{eq:exp3.1}\\
    \mathrm{ILOG2}(n)&= c - a / \mathrm{log}(n),\label{eq:ilog2}
\end{align}

where $n$ denotes the sample size (see Table~1 in ~\citet{viering2022shape} for further options).

The interpretation of the parameters in the \textit{EXP3} model in Eq.~\ref{eq:exp3.1} is as follows:
\vspace{-\topsep}
\begin{itemize}[noitemsep,topsep=0pt]
    \item[$n$]: sample size
    \item[$c$]: upper bound
    \item[$b$]: applicability threshold /\textit{a priori} sample size
    \item[$a$]: discounting factor for sample size expressing, e.g., noise
\end{itemize}

An important difference is that parameters $a$ and $b$ influence the learning rate, whereas parameter $c$ represents the asymptotic performance limit.

To model the contextual multi-task effects of all the other complementary tasks on a target task $i$ we use the following extension with two arguments
\begin{equation} \label{eq:exp3.2}
    \mathrm{EXP3.2}(n_i,n_\Sigma) = c_{i\Sigma} - \mathrm{exp}\left(b - a_i n_i-a_{i,\Sigma} n_\Sigma\right),
\end{equation}

where the extra argument $n_\Sigma$ is the sum of the sample sizes of the rest of the tasks
\begin{equation}\label{eq:gEXP3Sum}
    n_{\Sigma}=\left(\sum_{j=1, j\neq i}^K n_{j}\right)
\end{equation}

following \ref{eq:MTSampleSum}. Note that depending on the complementary set, this can model the effect of a specific group, a complete domain, and a general foundation. 

To model the focused pairwise effects of a specific auxiliary task $j$ we expand the model with a further argument and term using the sample size of an auxiliary task $n_j$
\begin{equation}\label{eq:exp3.3}
    \mathrm{EXP3.3}(n_i,n_\Sigma,n_j) = c_{ij} - \mathrm{exp}(b - a_i n_j-a_{ij} n_a-a_{i\Sigma} n_\Sigma).
\end{equation}

To standardize notation, we will denote the original single argument learning curve EXP3 as EXP3.1.

\section{Estimation of the MTLCs}

For a given data set with task sample sizes $\{n_i\}_{i=1\ldots K}$, the inductive performance and generalization error in Eq.~\ref{eq:IdealLearningCurve} can be approximated for any sample sizes $\{n'_i\leq n_i\}_{i=1\ldots K}$ by performing an MTL training and evaluation using all data subsets and complementary data set with appropriate sample sizes as suggested by Eq.~\ref{eq:IdealLearningCurve}. 
The \emph{statTAG} method offers an efficient method to generate grid data points and exploits an analogous advantage as the TAG algorithm for training curves~\citet{fifty2021efficiently}. Since our goal is to assess the contextual transfer effect between pairs of tasks, two kinds of data points are generated: (1) as reference points, the same set of folds are retained for all tasks and one multitask model is trained, for which performance metrics are calculated; (2) then, for each task $t_{aux}$, the former reference data set is augmented with a single further data fold, and the impact of $t_{aux}$ on the other tasks $t$ is assessed by comparing their newly achieved performance with the former ones.

In summary, three types of grid points are available in the estimation: (1) the single task learning grid points (STL); (2) the general multi-task learning grid points (MTL); and (3) the single task augmented multi-task learning grid points (statTAG). 

To increase the robustness of the estimation, we follow a three-stage process. At first, it estimates the parameters of the selected LC, notably EXP3.1 in Eq.\ref{eq:exp3.1} using the single task learning grid points. Secondly, it freezes the coefficient for the weight of the target task ($a_t$) and estimates EXP3.2 in Eq.\ref{eq:exp3.2} using the general multi-task learning grid points. Thirdly, it also freezes the coefficient for the weight of the complementary tasks ($a_\Sigma$) and estimates EXP3.3 in Eq.\ref{eq:exp3.3} using the single task augmented multi-task learning grid points. In summary, we fit the sample weight $a_t$ in EXP3.3(,0,0) using STL, then $a_\Sigma$ in EXP3.3(,,0) using standard MTL, and finally $a_t'$ for the pairwise auxiliary transfer effects using the statTAG data points.

\section{Materials and methods}

\subsection{Data set}

We use the KiBA data set, a high-quality kinase data set containing a special aggregated bioactivity score for $467$ targets and $52498$ KiBA compounds~\cite{tang2014making}. Of the $52498$ KiBA compounds, those lacking a smiles descriptor were discarded; thus, $52078$ were retained. SMILES descriptors were then standardized using the software tool RDKit, with the following results: $83$ compounds, which contained more than 100 non-H atoms, were discarded; among the remaining $51995$ canonical smiles descriptors, there were $50522$ unique ones, which then formed the final list of compounds, having the activation data corresponding to the original smiles descriptors merged. Since completely random train/test splits suffer from the compound series bias, leading to overoptimistic performance estimations~\cite{mayr18}, we utilize a more realistic, scaffold-based train/test split in the spirit of~\cite{simm21}. The scaffold to be assigned to the compound was selected in two steps: first, the Murcko scaffold belonging to the compound was retrieved using RDKit; second, this Murcko scaffold was used to select the preferred scaffold from the tree of scaffolds. Thus, the $50522$ unique canonical smiles descriptors were assigned to $9382$ scaffolds. Next, a random number in $[0, 9]$ was generated for each scaffold, defining a 10-fold cross-validation over the compounds. Finally, only those targets (columns of the DTI matrix) were kept, which had at least $100$ known KiBA-score values, thus resulting in a DTI matrix of the size $50522 \times 244$. Following DeepDTA~\cite{ozturk18}, we used a binary discretization scheme with a threshold of $pK_d=3$, resulting in $74779$ and $147470$ active and inactive DTIs, respectively. This data set is henceforth referred to as KIBA244. In evaluations, we fixed the ordering of the $10$ folds and used the $10$  rotated permutations in averaging. 

\subsection{Models and computations}

Our definition of the shared parameter model in Eq.~\ref{eq:CompositeMLP} follows the SparseChem multiple output MLP model with the decomposition at the penultimate layer as the $g$ foundation model and a ReLU activation in the $h$ for the task heads~\citet{arany2022sparsechem}. Trainings were performed using the SparseChem model, which consisted of a $32000$ neuron-wide input layer, a $500$ neuron-wide hidden layer, and an output layer the width of which was determined by the number of tasks present in the corresponding learning. ReLU and sigmoid activation functions were used in the first two layers and on the output. Hyperparameters for the training of neural networks were determined by grid search. During their training, ADAM optimization scheme was used with a learning rate of $0.0001$. Training was done for a fixed number of epochs, long enough to ensure convergence, depending on the complexity of data: $40$, $100$, and $300$ epochs, for trainings corresponding to single-task, within-group, and complete data settings. Neural network models were trained using the \emph{PyTorch} framework, utilizing $3$ \emph{Tesla V100} GPUs, each with $32$ GB memory and $1$ \emph{TITAN Xp} GPU with $12$ GB memory. The complexity of the trained models allowed $20$ and $10$ jobs to run in parallel on the Tesla V100 and the Titan XP GPUs, respectively, thus allowing for a parallelization of $70$ jobs in total. Running times were linear in the number of training epochs, with the longest, $300$ epoch-long jobs lasting for about $60$ minutes. The running time of supplementary calculations (e.g. calculation of the AUROC and AUPR curves for performance assessment) was negligible compared to the former.

\section{Results}

We present results about (1) the effect of multi-task learning; (2) the applicability of LCs, both in a static and online learning approach; (3) the characterization of learning dynamics in STL and MTL by LCs; (4) decomposition of global and pairwise transfer effects; (5) the applicability of gradient-based approaches to transfer in the DTI domain; (6) the relation of gradient-based and sample-based approaches to transfer; and (7) joint analysis of these two aspects of multi-task learning.

\subsection{Effect of multi-task learning}

\begin{figure}[ht]
    \centering
    \includegraphics[width=0.9\linewidth]{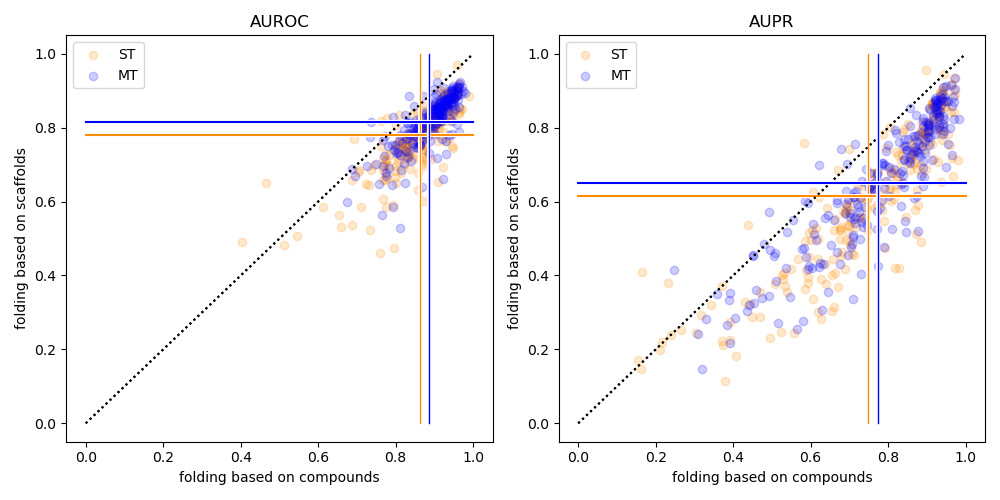}
    \caption{Comparison of the generalization performance estimations over the compounds versus over the scaffolds in the case of the KIBA data set using $1$ fold for training and the complementary data for testing.}
    \label{fig:CompoundVsScaffold_KIBA_fold9}
\end{figure}

Fig.~\ref{fig:CompoundVsScaffold_KIBA_fold9} shows the effect of using compounds instead of scaffolds in the case of the KIBA data set for $9$ folds for training and the complementary data for testing (results are averaged over the shifted fold permutations). Thus, we adopted this scheme in the paper.

We performed a systematic screening for the overall effect of multi-task learning comparing the performance of single-task and multi-task learning at various data sizes using $1,\ldots, 9$ folds in training.

\begin{figure}[ht]
\centering
\includegraphics[width=0.9\linewidth]{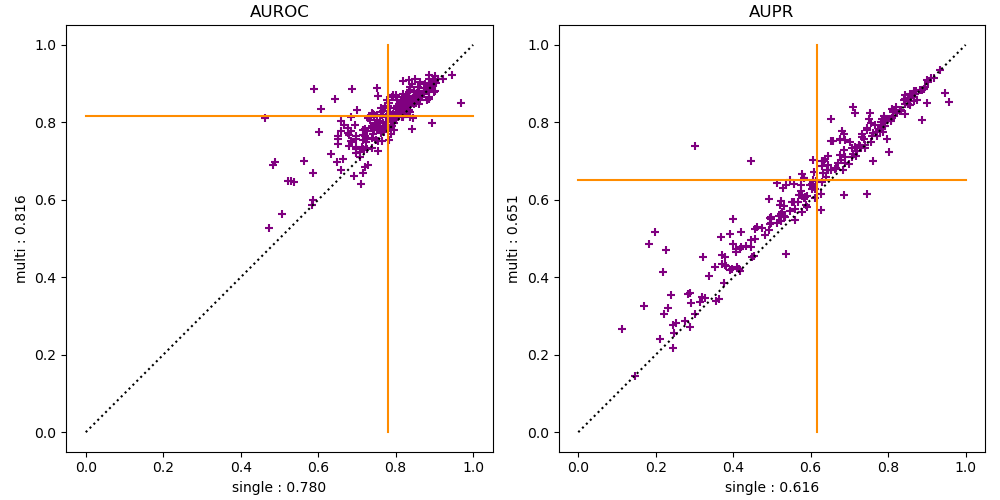} 
\label{fig:kiba_emp_scatter_scaff_fold9}
\caption{Task-by-task comparison of AUROC and AUPR performances in single-task and multi-task learning trained on $9$ folds of the KIBA244 data set, validated on the tenth fold; values averaged over the $10$ training-validation settings.}
\end{figure}

Overall, multi-task learning improved the AUROC performance of $137$ tasks at a nominally significant level and the AUPR performance of $94$ tasks.

\subsection{Selection of learning curves}

We compared the applicability of the functional forms of a wide range of learning curves, including the EXP4, EXP3, and ILOG2 models in Eq.~\ref{eq:exp4}, \ref{eq:exp3.1}, and \ref{eq:ilog2}.  We applied two performance measures in the evaluation: ($L_2$) the sum of squared errors of predictions for all data points and ('preq') its predictive sequential (prequential) variant, which sequentially fits the model for training data points corresponding to their respective folds, and evaluate its prediction for the next fold. To increase robustness, data points were averaged over the $10$ permutation shifts, denoted by an 'E' prefix.

\begin{table}
    \centering
    \begin{tabular}{r|cccc}
        ~ & $L_2$-9 & E[$L_2$-9] & preq & E[preq] \\ \hline
        ILOG2 & 2.34E-02 & 4.40E-04 & 2.28E-02 & 8.01E-04 \\ \hline
        \textbf{EXP3} & 2.30E-02 & 2.57E-04 & 2.29E-02 & 8.25E-04 \\ \hline
        EXP4 & 2.28E-02 & 1.83E-04 & 2.37E-02 & 1.56E-03 \\ \hline
    \end{tabular}
    \caption{Various error measures for LC fitting.}\label{t:LCSelection}
\end{table}

The EXP3 model exhibited superior or close to superior performance, so considering interpretability as well, we will use the EXP3.

\subsection{Learning curves of multitask learning}

\begin{figure}[ht]
    \centering
    \begin{subfigure}{0.5\textwidth}
    \includegraphics[width=0.9\linewidth]{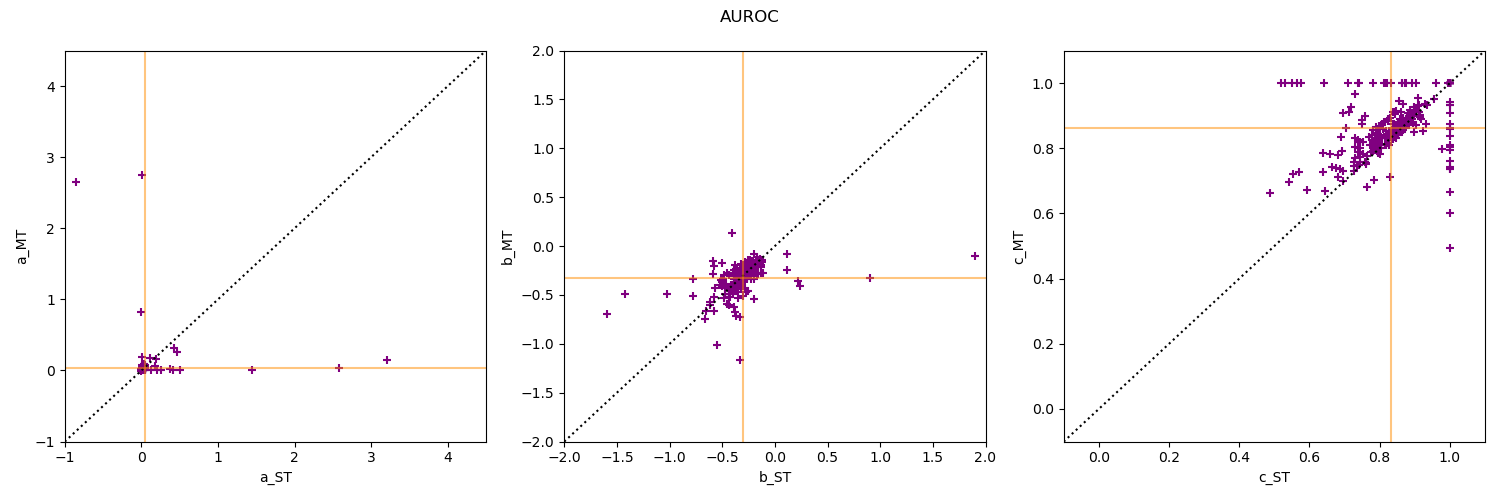} 
    \label{fig:scatter_params_ST-vs-MT__scaff_auroc}
    \end{subfigure}
    
    \begin{subfigure}{0.5\textwidth}
    \includegraphics[width=0.9\linewidth]{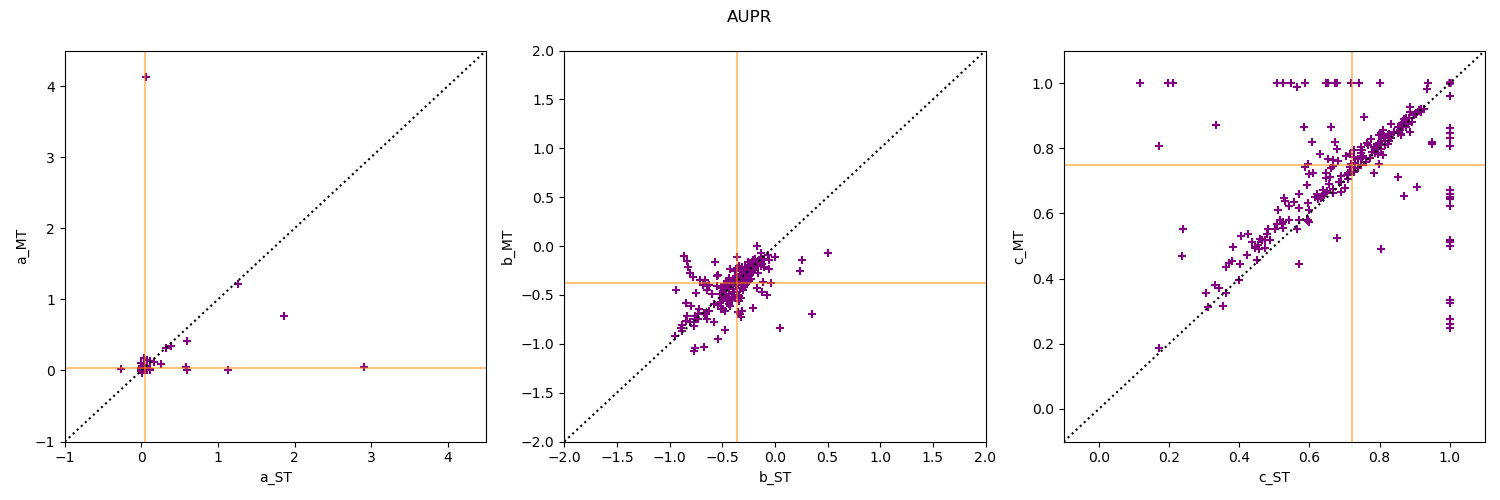}
    \label{fig:scatter_params_ST-vs-MT__scaff_aupr}
    \end{subfigure}
    
    \caption{Scatter plot showing the relation of EXP3.1 parameters between ST and MT learning curves for performance measures AUROC and AUPR.}
    \label{fig:scatter_params_ST-vs-MT__scaff}
\end{figure}

\begin{figure}
\centering
\includegraphics[width=0.9\linewidth]{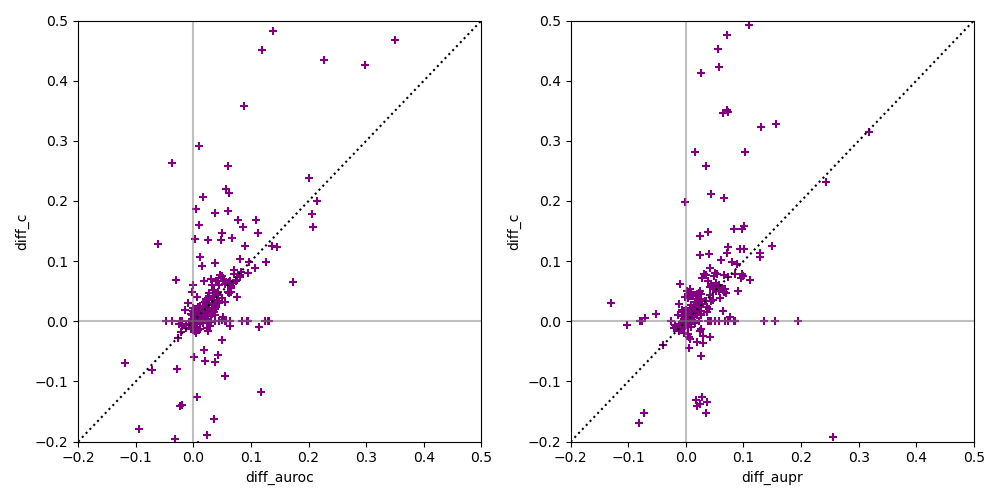} 
\caption{Scatter plot showing the relation between differences in performance measure AUROC and AUPR and differences in parameter $c_i$.}
\label{fig:kiba_scatter_comp__diff_meas_vs_param_c}
\end{figure}

The learning curves allow detailed characterizations of the inductive effects of multitask learning. Fig.~\ref{fig:scatter_params_ST-vs-MT__scaff} shows the comparison of the parameters $a_i,b_i,c_i$ of the EXP3.1 model in a single task versus multitask learning. Fig.~\ref{fig:kiba_scatter_comp__diff_meas_vs_param_c} shows the relation between the difference of performance measures and the difference of the EXP3.1 parameter $c_i$ describing asymptotic performances in STL vs MTL.

\begin{table}
    \centering
    \begin{tabular}{r|cc}
        ~ & AUROC & AUPR \\ 
        ~ & R (p-value) & R (p-value) \\ \hline
        a & -0.011 (8.624e-01) & -0.073 (2.601e-01) \\ \hline
        b & -0.171 (7.783e-03) & -0.160 (1.271e-02) \\ \hline
        c & 0.514 (1.048e-17) & 0.520 (3.680e-18) \\
    \end{tabular}
    \caption{Spearman rank correlation between differences of model parameters and differences of performance measures.}
    \label{tab:}
\end{table}

\subsection{Decomposition of transfer effects}

The multi-task extensions of learning curves, e.g., the EXP3.3 model, allow the delineation of domain-wide transfer effects versus the pairwise effects. Detailed results about the relation of model parameters and transfer effects are presented and discussed in the supplementary material.

\subsection{TAG}
In the study by \citet{fifty2021efficiently}, Task Affinity Groupings (TAG) was introduced as a method to determine the optimal clustering of tasks for training in multi-task neural networks. The primary objective of TAG is to improve overall efficiency by segmenting a wide range of tasks into more manageable subgroups. Within this framework, all tasks are integrated and trained within a singular model. A key feature of TAG is its assessment of the influence of parameter updates for a specific task on the losses associated with other tasks, a process termed inter-task affinity. Drawing inspiration from meta-learning, TAG employs a comparable approach by adjusting parameters for individual tasks and subsequently examining their repercussions on other tasks. By gathering data on task interrelations, TAG distinguishes between tasks that exhibit positive and negative interactions. This discerned knowledge facilitates the formation of task groupings that amplify inter-task affinity, which in turn augments model performance.

Despite utilizing TAG for the KIBA244 data set, we did not succeed in identifying a task grouping that outperformed the traditional multitask setting reliably. The details of our experiments are provided in the supplementary materials. We also explored the efficacy of GradNorm \citet{chen2018gradnorm}, an algorithm designed to balance training in deep multitask networks, and found it had no effect on model performance on the KiBA244. The details of our experiments are provided in the supplementary materials.

\subsection{TAG vs STAG}

\begin{figure}
\centering
\includegraphics[width=1.0\linewidth]{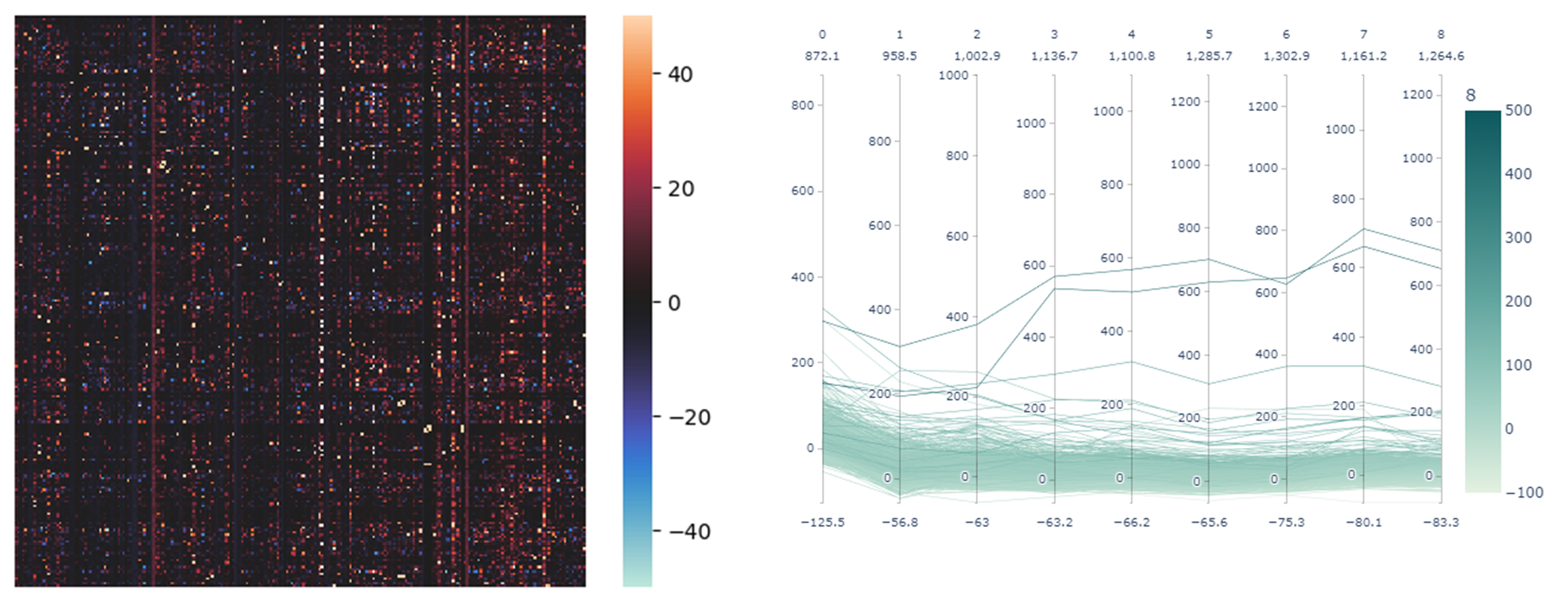} 
\caption{Heatmap representation of TAG scores derived from all nine folds, accompanied by parallel coordinates illustrating the progression of TAG scores with increasing amounts of training data.}
\label{fig:TAG_MERGED}
\end{figure}

To explore the relation between gradient-based approaches, such as TAG and GradNorm, and the MTLCs we used the following analogous study design for varying sample sizes. First, we expanded TAG to collect information about the domain-wide transfer effects of the complementary tasks, as it was neglected in the original work. Next, we estimated both this domain-wide transfer effect and the pairwise effects in TAG for an increasing number of folds and shifted permutations of the KIBA244 data set. Finally, we compared these domain-wide and pairwise TAG estimates to analogous descriptors of transfer effects in the EXP3.3 model, notably to the sample weights $a_{i,\Sigma},a_{i,j}$ and asymptotic performance coefficients $c_{i,j}$.  Fig.~\ref{fig:TAG_MERGED} illustrates the pairwise transfer effects in TAG. The Spearman rank correlations between transfer effect coefficients in TAG and EXP3.3 are summarized in Table~\ref{tab:X2DO}. See the supplement for illustrations of the coefficients for the pairwise transfer effects in the EXP3.3 model and further comparisons. 

\begin{table}
    \centering
    \begin{tabular}{r|l|l}
        ~ & 1-fold & averaged \\ \hline
        a\_auroc & 0.037 (1.283e-17) & 0.064 (2.371e-51) \\ \hline
        a\_aupr & 0.023 (8.551e-08) & 0.060 (1.672e-44) \\ \hline
        b\_auroc & 0.001 (8.386e-01) & -0.004 (3.940e-01) \\ \hline
        b\_aupr & 0.008 (7.042e-02) & 0.008 (5.931e-02) \\ \hline
        c\_auroc & 0.059 (4.655e-43) & 0.031 (2.145e-13) \\ \hline
        c\_aupr & 0.258 (0.000e+00) & 0.201 (0.000e+00) \\
    \end{tabular}
    \caption{Spearman rank correlations between transfer effect coefficients in TAG and the ones  derived from the value ($a_{i,j}$) or the difference ($b_{i,j}-b_{i,\Sigma}$ and $c_{i,j}-c_{i,\Sigma}$) of different parameters of EXP3.3.}
    \label{tab:X2DO}
\end{table}

\section{Discussion}

Transfer effects in multi-task learning still pose theoretical and practical challenges, and the DTI prediction problem provides a unique domain with fundamental relevance and industry-scale data sets. Our results related to domain-wide effects ($a_{i,\Sigma}$) confirm that the foundation layer in the kinome domain with 450 tasks does transfer general information about the unified ligand-target chemogenomic space. Indeed, the proposed MTLCs indicated multiple aspects of significant domain-wide --- foundation or deep --- transfer. Our results are also a demonstration that the proposed MTLC models can indicate and quantitatively characterize transfer for incomplete and missing-not-at-random data in a notoriously challenging domain of DTI prediction. The explored parallels related to MTL transfer between training curves (optimization) and learning curves (induction) also suggest large-scale orchestration of training tasks in MTL, such as in curriculum learning~\citet{bengio2009curriculum,pentina2015curriculum}. Finally, the MTLCs are the critical ingredients in multi-task active learning; thus, both their theoretical underpinnings, the improved estimation using reinforcement learning strategies in the grid point generation and estimation, and the inference of a global model from the estimated marginal MTLCs are promising research directions.

Our current work aimed at improving the detection of transfer, but improved estimates and characterizations of transfer effects are not used to improve the MTL performance for a fixed task set and data set. However, information about task saturation can be used to scale the gradients or virtual sample size in MTL; improved transfer estimates can lead to better task decomposition; and MTLCs can be used in budgeted learning to select the most useful experiments. 

\section{Conclusion}

This paper proposes multi-task learning curve (MTLC) approximations to quantitatively model transfer effects and an efficient grid data point generation method. Systematic evaluations of the MTLCs in a benchmark drug-target interaction data set confirmed that they could detect and predict performance in a wide range of sample sizes, despite the fact that existing gradient orchestration methods, such as GradNorm, and multi-task grouping methods, such as TAG and MTG-Net could not improve performance. The significantly higher computational cost of utilizing learning curves over samples and not batches in the training process allowed the quantitative modeling of transfer effects in this previously inaccessible area of drug-target interaction prediction. Our work hopefully contributes to a better understanding of transfer in areas with highly incomplete and missing-not-at-random data sets, provides an interesting parallel in two aspects of learning, and helps the development of provably approximately correct analysis of foundation models using domain-wide or universal representations. 

\section{Acknowledgments}
This research was supported by the National Research, Development, and Innovation Fund of Hungary under Grant OTKA-K139330, TKP2021-EGA-02, and the European Union project RRF-2.3.1-21-2022-00004 within the framework of the Artificial Intelligence National Laboratory. This research was carried out as part of E-GROUP ICT SOFTWARE Zrt.'s “IPCEI-CIS FedEU.ai – Federated Cloud-Edge AI” project, supported by the Government of Hungary and the Ministry of National Economy, under grant number NGM/4043/1/2024. The authors acknowledge the support of E-GROUP ICT SOFTWARE Zrt. for providing the research infrastructure and resources essential to this work. The research presented in this paper forms part of E-GROUP’s core research domain and contributes to the advancement of its strategic directions in federated, cloud-edge, and cutting-edge AI technologies. This study was also supported by the European Union project RRF-2.3.1-21-2022-00004 within the framework of the Artificial Intelligence National Laboratory.


\end{document}